# Submodular Decomposition Framework for Inference in Associative Markov Networks with Global Constraints

Anton Osokin[*]   Dmitry Vetrov[*]

Department of Computational Mathematics
and Cybernetics, Moscow State University
anton.osokin@gmail.com   vetrovd@yandex.ru

Vladimir Kolmogorov

Department of Computer Science
University College London, UK
v.kolmogorov@cs.ucl.ac.uk

## Abstract

In the paper we address the problem of finding the most probable state of discrete Markov random field (MRF) with associative pairwise terms. Although of practical importance, this problem is known to be NP-hard in general. We propose a new type of MRF decomposition, submodular decomposition (SMD). Unlike existing decomposition approaches SMD decomposes the initial problem into subproblems corresponding to a specific class label while preserving the graph structure of each subproblem. Such decomposition enables us to take into account several types of global constraints in an efficient manner. We study theoretical properties of the proposed approach and demonstrate its applicability on a number of problems.

## 1 Introduction

The problem of effective Bayesian inference arises in many applied domains, e.g. in machine learning, computer vision, decision-making, etc. One of the most intriguing problems is the development of approximate inference algorithms for the problems that are NP-hard in general. An important particular case is the inference problem in cyclic discrete Markov random fields (MRF) with energies that can be represented via sum of unary and pairwise terms.

Let $G = (\mathcal{V}, \mathcal{E})$ be an undirected graph with $\mathcal{V}$ and $\mathcal{E}$ being the sets of nodes and edges respectively. With each node we associate a class label $t_j \in \{1, \ldots, P\}$. The inference problem can be formulated as an energy minimization problem

$$\sum_{(j \in \mathcal{V})} \theta_j(t_j) + \sum_{(i,j) \in \mathcal{E}} \theta_{ij}(t_i, t_j) \to \min_{t_1, \ldots, t_{|\mathcal{V}|} \in \{1, \ldots, P\}}. \tag{1}$$

where $\theta_j(t_j)$ and $\theta_{ij}(t_i, t_j)$ are some known functions of discrete argument.

Although NP-hard in general there are several special cases when the inference problems can be solved exactly in polynomial time. One example is dynamic programming approach [21] for inference in tree-structured graphs. Another example are MRFs with outer-planar graphs [23], or more generally MRFs with low treewidth [19]. Min-cut/max-flow algorithms can efficiently solve the inference problem on arbitrary graphs when all the variables are binary and pairwise potentials meet submodularity constraints [8, 15].

$$\theta_{ij}(0,0) + \theta_{ij}(1,1) \leq \theta_{ij}(0,1) + \theta_{ij}(1,0). \tag{2}$$

The multi-class problems on cyclic graphs do not have efficient exact algorithms except few very special cases with specific orderings in the space of classes [9, 6].

Recently advanced approximate methods based on MRF decomposition have appeared [24]. The most popular method tree-reweighted message passing (TRW) [29, 13, 17] splits the MRF with cycles into a number of acyclic subgraphs (trees), for each of them inference is made independently with the subsequent harmonization of optimal solutions. The other decomposition methods (e.g. [30, 25, 16, 1]) exploit similar ideas. Unlike more efficient approximate energy minimization algorithms, e.g. [5, 18], decomposition framework makes it possible to take into account some global properties of the solution, e.g. to establish constraints on the areas of classes [31, 32, 20].

---
[*]The authors assert equal contribution and thus joint first authorship.

In this paper we develop an alternative framework for approximate inference in MRF with associative pairwise terms. We present submodular decomposition (SMD) technique. Instead of subgraph-based decomposition we split the problem into a number of submodular problems keeping the structure of graph $\mathcal{G}$. To harmonize optimal solutions of different subproblems we use the dual decomposition approach [2]. We show that SMD provides energy lower bounds which are equivalent to the ones by TRW. However the new type of decomposition allows us to take into account the preferences on any type of global linear statistics of the class indicator variables in a straightforward manner as well as to establish some shape constraints.

The rest of paper is organized as follows. In the next section we present SMD framework. We prove the equivalence of SMD and TRW lower bounds in section 3 and examine the properties of SMD convergent point in section 4. Section 5 presents a possible way for optimizing the lower bound which is based on the averaging of min-marginals. The different types of global constraints which can be implemented efficiently into SMD framework are discussed in section 6. We present the experimental evaluation of our method in section 7 and give some conclusions in section 8.

## 2 Submodular Decomposition

Consider the indicator parametrization of (1) obtained by establishing auxiliary binary variables $Y = \{y_{jp}\} \in \{0,1\}^{|\mathcal{V}| \times P}$:

$$y_{jp} = \begin{cases} 1, & t_j = p, \\ 0, & \text{otherwise.} \end{cases}$$

Denote $\theta_j(p) = \theta_{jp}$ and $\theta_{ij}(p,q) = \theta_{ij,pq} = \theta_{ji,qp}$. Then the problem (1) takes the form

$$E(Y) = \sum_{j \in \mathcal{V}} \sum_{p=1}^{P} \theta_{jp} y_{jp} + \sum_{(i,j) \in \mathcal{E}} \sum_{p,q=1}^{P} \theta_{ij,pq} y_{ip} y_{jq} \to \min_{Y \in \mathcal{L} \cap \mathcal{G}}. \tag{3}$$

Here we denoted the set of *local* constraints by $\mathcal{L} = \{Y \mid y_{jp} \in \{0,1\}\}$ and the set of *global* constraints by $\mathcal{G} = \{Y \mid \sum_{p=1}^{P} y_{jp} = 1, \forall j \in \mathcal{V}\}$. The first group of constraints requires $y_{jp}$ to be indicator variables while the second group provides consistency of graph labeling, i.e. each node belongs to one and only one class.

In what follows we assume the associativeness [26] of pairwise terms, i.e.

$$\theta_{ij,pq} = -C_{ij,p} \delta_{pq}, \tag{4}$$

where $\delta_{pq} = 1$ iff $p = q$ and $C_{ij,p}$ are non-negative constants[1]. Then we may rewrite the energy (3) as follows

$$\min_{Y \in \mathcal{L} \cap \mathcal{G}} E(Y) =$$

$$\min_{Y \in \mathcal{L} \cap \mathcal{G}} \sum_{p=1}^{P} \left( \sum_{j \in \mathcal{V}} \left( \theta_{jp} - \frac{1}{2} \sum_{i:(i,j) \in \mathcal{E}} C_{ij,p} \right) y_{jp} + \frac{1}{2} \sum_{(i,j) \in \mathcal{E}} C_{ij,p} \left[ y_{ip}(1 - y_{jp}) + y_{jp}(1 - y_{ip}) \right] \right) =$$

$$\min_{Y_p \in \{0,1\}^{|\mathcal{V}|}, \sum Y_p = \vec{1}} \sum_{p=1}^{P} E_p(Y_p) \geq \min_{Y_p \in \{0,1\}^{|\mathcal{V}|}} \sum_{p=1}^{P} E_p(Y_p) = \sum_{p=1}^{P} \min_{Y_p \in \{0,1\}^{|\mathcal{V}|}} E_p(Y_p). \tag{5}$$

---
[1]Note that by setting all $C_{ij,p} = C_{ij}$ we get an important case of generalized Potts potential.

Here we denoted the $p^{th}$ column of $Y$ by $Y_p$ and $(i,j)$ is an undirected edge of the graph. Note that each $E_p(Y_p)$ is a submodular function so that we can find $\min_{Y_p \in \{0,1\}^{|\mathcal{V}|}} E_p(Y_p)$ efficiently by min-cut/max-flow algorithm (e.g. [4]). To reconcile the minimizers of $E_p$ consider a Lagrangian

$$L(Y, \Lambda) = \sum_{p=1}^{P} E_p(Y_p) + \sum_{j \in \mathcal{V}} \lambda_j \left( \sum_{p=1}^{P} y_{jp} - 1 \right). \quad (6)$$

The following lower bound is valid for the initial problem (3):

$$\min_{Y \in \mathcal{L} \cap \mathcal{G}} E(Y) = \min_{Y \in \mathcal{L} \cap \mathcal{G}} L(Y, \Lambda) \geq \max_{\Lambda} \min_{Y \in \mathcal{L}} L(Y, \Lambda) =$$

$$= \max_{\Lambda} \left[ \sum_{p=1}^{P} \min_{Y_p \in \{0,1\}^{|\mathcal{V}|}} \left( E_p(Y_p) + \sum_{j \in \mathcal{V}} \lambda_j y_{jp} \right) - \sum_{j \in \mathcal{V}} \lambda_j \right]. \quad (7)$$

Note that the minimization of

$$\Phi_p(Y_p, \Lambda) = E_p(Y_p) + \sum_{j \in \mathcal{V}} \lambda_j y_{jp}$$

w.r.t. $Y_p$ can still be done by min-cut algorithms since the functions remain submodular. The maximization problem (7) is concave w.r.t. $\Lambda$ and hence the solution can be found via subgradient ascent. Thereby we replace NP-hard optimization problem (3) with a series of simpler problems that can be solved efficiently. Hereinafter we refer to both decomposition and algorithm as submodular decomposition (SMD).

Note, that $\min_{Y \in \mathcal{L} \cap \mathcal{G}} E(Y) = \min_{Y \in \mathcal{L}} \max_\Lambda L(Y, \Lambda)$ and energy lower bound (6) equals the maximin value of Lagrangian $\max_\Lambda \min_{Y \in \mathcal{L}} L(Y, \Lambda)$. Now consider the continuous relaxation $\mathcal{Q} = \{Y | y_{jp} \in [0,1]\}$ of local constraints. It can be shown [22, Theorem 3.1] that $\min_{Y \in \mathcal{L} \cap \mathcal{G}} E(Y) = \min_{Y \in \mathcal{Q} \cap \mathcal{G}} E(Y)$ for arbitrary number of classes, unary and pairwise potentials, hence

$$\min_{Y_p \in \{0,1\}^\mathcal{V}} \Phi_p(Y_p, \Lambda) = \min_{Y_p \in [0,1]^\mathcal{V}} \Phi_p(Y_p, \Lambda).$$

This implies

$$L^* = \min_{Y \in \mathcal{L} \cap \mathcal{G}} E(Y) = \min_{Y \in \mathcal{Q}} \max_\lambda L(Y, \Lambda) \geq \max_\Lambda \min_{Y \in \mathcal{L}} L(Y, \Lambda) = \max_\Lambda \min_{Y \in \mathcal{Q}} L(Y, \Lambda) = L_*. \quad (8)$$

In the space $(\mathcal{Q} \cap \mathcal{G}) \times \mathbb{R}^{|\mathcal{V}|}$ the Lagrangian (6) is smooth and differentiable function, hence we can find the global minimum of energy (3) by performing SMD if and only if $L(Y, \Lambda)$ has a saddle-point in $(\mathcal{Q} \cap \mathcal{G}) \times \mathbb{R}^{|\mathcal{V}|}$.

## 3 Equivalence of Lower Bounds

In this section we show that the optimal value of Lagrangian dual (7), which is solved by SMD, is equivalent Kleinberg-Tardos (KT) relaxation [10] of uniform metric labeling problem. Furthermore we prove that KT-relaxation is equivalent to Schlesinger LP-relaxation which is reviewed in [30] and is solved e.g. by TRW [29].

**Theorem 1** *The maximin value the Lagrangian (6) equals the minimum value of KT-relaxed problem [10]:*

$$L_* = \min_X \sum_{j \in \mathcal{V}} \sum_{p=1}^{P} \theta_{jp} x_{jp} + \frac{1}{2} \sum_{(i,j) \in \mathcal{E}} \sum_{p=1}^{P} C_{ij,p} x_{ij,p}, \quad (9)$$

$$s.t. \sum_{p=1}^{P} x_{ip} = 1,$$

$$x_{ij,p} \geq x_{ip} - x_{jp}, \quad x_{ij,p} \geq x_{jp} - x_{ip},$$

$$x_{ij,p} \geq 0, \quad x_{ip} \geq 0.$$

See the proof in appendix A.

**Theorem 2** *If energy* (3) *satisfies the condition* (4) *then the KT-relaxation [10] is equivalent to the Shlesinger LP-relaxation:*

$$L_* = \min_Z \sum_{j \in \mathcal{V}} \sum_{p=1}^P \theta_{jp} z_{jp} + \sum_{(i,j) \in \mathcal{E}} \sum_{p,q=1}^P d_{ij,pq} z_{ij,pq}, \quad (10)$$

$$s.t. \sum_{p=1}^P z_{ip} = 1,$$

$$\sum_{p=1}^P z_{ij,pq} = z_{jq}, \quad \sum_{q=1}^P z_{ij,pq} = z_{ip},$$

$$z_{ij,pq} \geq 0, \quad z_{ip} \geq 0,$$

where $d_{ij,pq} = 0$ iff $p = q$ and $\frac{C_{ij,p} + C_{ij,q}}{2}$ otherwise.

See the proof in appendix B.

## 4 Properties of the Maximin Point

Consider the properties of $L(Y, \Lambda)$ when $Y \in \mathcal{L}$, i.e. all $y_{jp} \in \{0, 1\}$. Then the maximin problem (7) can be viewed as maximization of a piecewise linear function $\min_{Y \in \mathcal{L}} L(Y, \Lambda)$ in the space of $\Lambda$'s. Each $Y \in \mathcal{L}$ defines a hyperplane in this space and $\Omega = \{(l, \Lambda) \mid l \leq \min_{Y \in \mathcal{L}} L(Y, \Lambda)\}$ is a concave polytope.

**Lemma 1** *Let $(Y^0, \Lambda^0)$ be a maximin point of the Lagrangian, i.e. $L(Y^0, \Lambda^0) = \max_\Lambda \min_{Y \in \mathcal{L}} L(Y, \Lambda)$. The global minimum of energy $E(Y)$ equals $L(Y^0, \Lambda^0)$ if and only if there is a horizontal hyperplane defined by $Y^1$ such that $L(Y^1, \Lambda^0) = L(Y^0, \Lambda^0)$.*

*If it is known additionally that subgradient of $\min_{Y \in \mathcal{L}} L(Y, \Lambda)$ at the point $\Lambda^0$ consists of only one vector then $Y^0$ is the optimal solution of* (3).

See the proof in appendix C.

Since $\min_{Y \in \mathcal{L}} L(Y, \Lambda)$ is a piecewise linear function three cases are possible at its maximum (see fig. 1). In the first case there is a plateau on the top of the polytope. In the second case there is no plateau but there exists a horizontal hyperplane defined by $Y^1$ that touches polytope $\Omega$. The third case corresponds to the non-zero duality gap. Note that in the first case we get the optimal value of $Y^0 = \arg\min_{Y \in \mathcal{L} \cap \mathcal{G}} E(Y)$ whereas in the second case in general we may not get the optimal value of $Y^1 \in \mathcal{G}$ in an explicit manner. Several cases when it becomes possible are considered further.

Now we define and investigate strong agreement (SA) and weak agreement (WA) conditions that are analogous to strong tree agreement (STA) and weak tree agreement (WTA)[2].

Denote the possible optimal labelings of the node $j$ in subproblem $p$ given $\Lambda$ as

$$Z_{jp}(\Lambda) = \{z \mid \exists Y_p^* = (y_{1p}^*, \ldots, y_{|\mathcal{V}|p}^*) : y_{jp}^* = z, \ \Phi_p(Y_p^*, \Lambda) = \min \Phi_p(Y_p, \Lambda)\}.$$

**Definition 1** *The variables $\Lambda$ satisfy strong agreement condition if*

$$\forall j \ \exists! p : \ Z_{jp}(\Lambda) = \{1\}, \ \forall q \neq p : \ Z_{jq}(\Lambda) = \{0\}.$$

From the last definition it follows that $\Lambda^0$ satisfies strong agreement if and only if $\arg\min_{Y \in \mathcal{L}} L(Y, \Lambda^0) \in \mathcal{G}$.

**Theorem 3** *If there exists $\Lambda^0$ that satisfies strong agreement condition and $Y^0 = \arg\min_{Y \in \mathcal{L}} L(Y, \Lambda^0)$, then $E(Y^0) = \min_{Y \in \mathcal{L} \cap \mathcal{G}} E(Y)$.*

---
[2]WTA and STA were defined in [14] for TRW framework. In this paper we explore similar conditions for SMD.

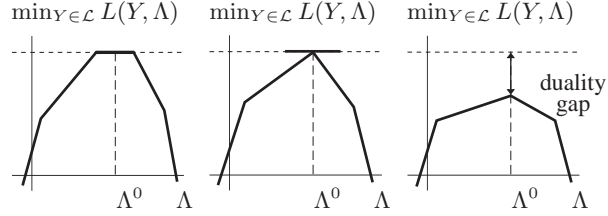

Figure 1: Three possible cases in SMD dual problem. The optimal energy value is shown by dotted line. The left plot corresponds to the case when we are able to find both optimal energy and optimal configuration of labels. The middle plot shows the case when there is zero duality gap but we are not guaranteed to find the optimal configuration of labels. The horizontal face corresponds to $Y^1$ (see lemma 1). The right plot illustrates the case when there's no horizontal face coming through the maximin point. Then we observe a non-zero duality gap.

**Definition 2** *The variables $\Lambda$ satisfy weak agreement condition if for all $j$ both statements hold true:*

1. $\exists q : 1 \in Z_{jq}(\Lambda)$.

2. *If* $\exists p : Z_{jp}(\Lambda) = \{1\}$, *then* $\forall q \neq p \Rightarrow 0 \in Z_{jq}(\Lambda)$;

It is easy to see that strong agreement condition implies weak agreement condition.

**Theorem 4** *If the point $(\Lambda^0, Y^0)$ is a maximin point of $L(Y, \Lambda)$ then variables $\Lambda^0$ satisfy weak agreement condition.*

**Proof.** Assume the contrary. Let $(\Lambda^0, Y^0)$ be a maximin point. There are two possible cases: there exist $j$, $p$, and $q$ such that $Z_{jp}(\Lambda^0) = Z_{jq}(\Lambda^0) = \{1\}$; there exists $j$, such that for all $p$ it holds $Z_{jp}(\Lambda^0) = \{0\}$. In the first case we may always take $\Lambda^1$ such that for all $i \neq j$ holds $\lambda_i^1 = \lambda_i^0$ and $\lambda_j^1 = \lambda_j^0 + \varepsilon$, where $\varepsilon > 0$. If $\varepsilon$ is small enough $\{1\}$ still belongs to both $Z_{jp}(\Lambda^1)$ and $Z_{jq}(\Lambda^1)$. Hence $Y^0 \in \text{Arg min } L(Y, \Lambda^1)$ but $L(Y^0, \Lambda^1) > L(Y^0, \Lambda^0)$ that contradicts the condition that $(\Lambda^0, Y^0)$ is maximin point of the Lagrangian. The second case is considered in a similar manner. ∎

Let $Free(p, \Lambda)$ be a set of nodes such that $Z_{jp}(\Lambda) = \{0, 1\}$, and let $Free^k(p, \Lambda)$ be its $k^{th}$ connected component.

**Theorem 5** *Let $Y_p^* = \arg\min_{Y_p} \Phi_p(Y_p, \Lambda)$. Consider $Y_p^{(1)}$ such that*

$$y_{jp}^{(1)} = \begin{cases} 1, & j \in Free(p, \Lambda), \\ y_{jp}^*, & \text{otherwise}; \end{cases}$$

*and $Y_p^{(0)}$ such that*

$$y_{jp}^{(1)} = \begin{cases} 0, & j \in Free(p, \Lambda), \\ y_{jp}^*, & \text{otherwise}. \end{cases}$$

*Then the following is true*

$$\Phi_p(Y_p^{(1)}, \Lambda) = \Phi_p(Y_p^{(0)}, \Lambda) = \Phi_p(Y_p^*, \Lambda).$$

The proof follows from submodularity inequality for the function $E_p(Y_p)$.

Note that the last theorem can be trivially generalized to separate connected components. In each $Free^k(p, \Lambda)$ we may replace $y_{jp}^*$ with either all ones or all zeros regardless of other $Free^l(p, \Lambda)$ and still keep $\Phi_p(Y_p, \Lambda)$ at its minimum. This is an important property since it helps to harmonize the solutions of subproblems.

**Theorem 6** *Assume that $\Lambda^0$ satisfies weak agreement condition. Then for any $\varepsilon > 0$ there always exists $\Lambda^1$ such that:*

- $||\Lambda^0 - \Lambda^1|| < \varepsilon$;

- $\Lambda^1$ *also satisfies weak agreement condition;*

- $\min_{Y \in \mathcal{L}} L(Y, \Lambda^0) = \min_{Y \in \mathcal{L}} L(Y, \Lambda^1)$;

- *If for any $j$ there is a subproblem $p$ such that $j \in Free(p, \Lambda^1)$ then*

    - *there exists at least yet another subproblem $q$ such that $j \in Free(q, \Lambda^1)$;*
    - *for all problems $r$ such that $j \notin Free(r, \Lambda^1)$ it follows $Z_{jr}(\Lambda^1) = \{0\}$.*

**Proof.** We provide here a sketch of the proof.

At the point $\Lambda^0$ the Lagrangian has either a plateau or a peak w.r.t. each variable $\lambda_j$. In case of plateau if $\lambda_j$ is on the border then take $\lambda_j \pm \varepsilon$ in order to be *inside* the plateau. At each interior point of the plateau $Z_{jp}(\Lambda^0)$ consists of a single element for each $p$.

If there is no plateau then $\frac{\partial L(Y,\Lambda)}{\partial \lambda_j}$ has a jump of at least 2 at the point $\Lambda^0$. This means that there are at least two subproblems $p$ and $q$ such that $Z_{jp}(\Lambda^0) = Z_{jq}(\Lambda^0) = \{0, 1\}$.

If for $\Lambda^0$ there exists $r$ such that $Z_{jr} = \{1\}$ then set $\lambda_j^1 = \lambda_j^0 + \varepsilon$. $Z_{jr}(\Lambda^1)$ still contains $\{1\}$ while for all other subproblems $Z_{jq}(\Lambda^1)$ consists of only zeros. ∎

Without loss of generality we may consider only those $\Lambda^1$ that satisfy the last theorem. Then the following theorems hold.

**Theorem 7** *Assume that $\Lambda^0$ satisfies weak agreement condition and $Y^0 = \arg\min_{Y \in \mathcal{L}} L(Y, \Lambda^0)$. If there exists such subproblem $p$ that for all $q$ it follows that $Free(q, \Lambda^0) \subseteq Free(p, \Lambda^0)$ then setting*

$$y^*_{jr} = \begin{cases} 1, & j \in Free(r, \Lambda^0), \ r = p; \\ 0, & j \in Free(r, \Lambda^0), \ r \neq p; \\ y^0_{jr}, & \text{otherwise}, \end{cases}$$

*we get an optimal configuration $Y^* = \arg\min_{Y \in \mathcal{L} \cap \mathcal{G}} E(Y)$.*

This theorem is a corollary of theorems 5 and 6. Note that it can be straightforwardly extended to the case of separate connected components.

**Theorem 8** *Assume that $\Lambda^0$ satisfies weak agreement condition and $Y^0 = \arg\min_{Y \in \mathcal{L}} L(Y, \Lambda^0)$. If there exists a partition $Free^{k_1}(p_1, \Lambda^0), \ldots, Free^{k_m}(p_m, \Lambda^0)$ of the set $I = \{j | \exists p : Z_{jp} = \{0, 1\}\}$ such that $Free^{k_i}(p_i, \Lambda^0) \cap Free^{k_j}(p_j, \Lambda^0) = \emptyset$ and $I = \bigcup_i Free^{k_i}(p_i, \Lambda^0)$ then an optimal configuration $Y^* = \arg\min_{Y \in \mathcal{L} \cap \mathcal{G}} E(Y)$ can be constructed as follows*

$$y^*_{jr} = \begin{cases} y^0_{jr}, & j \notin I; \\ 1, & r = p_i, \ j \in Free^{k_i}(p_i, \Lambda^0), \ \forall i = 1, \ldots, m; \\ 0, & \text{otherwise}. \end{cases}$$

**Proof.** For each $i$ we apply theorem 7 to $Free^{k_i}(p_i, \Lambda^0)$. ∎

The last theorem allows us to get an optimal solution of initial problem in the some cases when duality gap is zero but there is no plateau in the top of a polytope defined by $\min_{Y \in \mathcal{L}} L(Y, \Lambda)$ (case 2 in lemma 1).

## 5 Min-Marginals Averaging

In this section we derive an alternative approach to the optimization of SMD lower bound based on the min-marginals averaging. First we prove an important lemma.

Suppose that for each of our subproblems we have estimated unary min-marginals $\hat{\theta}_{jp}(k, \Lambda) = \min_{\{Y_p | y_{jp}=k\}} \Phi_p(Y_p, \Lambda)$. This can be done effectively e.g. using dynamic graphcut framework [11]. Now consider the following re-estimation equations for Lagrange multipliers

$$\lambda_j^{new} = \lambda_j^{old} + \Delta\lambda_j, \tag{11}$$

where $\Delta\lambda_j = \max_p \left[ \hat{\theta}_{jp}(0, \Lambda^{old}) - \hat{\theta}_{jp}(1, \Lambda^{old}) \right]$.

**Theorem 9** *The iterative process defined by* (11) *has the following properties:*

- *The lower bound of the Lagrangian does not decrease at each step;*
- *The fixed point of the process satisfies weak agreement condition.*

The proof is in appendix D.

**Theorem 10** *If $\Delta\lambda_j^0 = 0$ for some point $\Lambda^0$ then $\Lambda^0$ is a maximum point of $L(Y, \Lambda)$ w.r.t. $\lambda_j$ given all other $\lambda$'s are fixed.*

**Proof.** The condition $\Delta\lambda_j = 0$ implies $\max_p(\hat{\theta}_{jp}(0) - \hat{\theta}_{jp}(1)) = 0$. If we increase $\lambda_j$ all $y_{jp}$ are assigned the value of 0 and hence the Lagrangian decreases. Similarly the decrease of $\lambda_j$ implies the decrease of the Lagrangian since there is at least one subproblem $p$ for which $y_{jp} = 1$. ∎

From the last theorem it follows that unless $\Lambda^0$ satisfies weak agreement condition, among the nodes with $\Delta\lambda_j \neq 0$ there always exists at least one node $i$ such that changing $\lambda_i$ by $\Delta\lambda_i$ will definitely increase the value of the Lagrangian. The question of how to find such $i$ is easy to answer. Any $\Delta\lambda_j < 0$ leads to the increase of $L(Y, \Lambda)$. If all $\Delta\lambda_j \geq 0$ then we should select $i$ for which there exist at least two subproblems $p$ and $q$ such that

$$\begin{cases} \hat{\theta}_{ip}(0, \Lambda) > \hat{\theta}_{ip}(1, \Lambda) \\ \hat{\theta}_{iq}(0, \Lambda) > \hat{\theta}_{iq}(1, \Lambda). \end{cases}$$

The update of such $\lambda_j$ increases the value of the Lagrangian[3]. However we found the process of min-marginals averaging to be too slow in practice even with the use of dynamic graphcuts [12]. So in our implementation we mostly used subgradient ascend switching to min-marginals averaging for one iteration when the lower bound started oscillating.

The possibility of min-marginals averaging and weak agreement condition shows that there are lots of similarities between various types of decomposition as well as optimization on trees and optimization of submodular functions on cycled graphs have much in common.

## 6 The Inclusion of Global Terms

The submodular decomposition allows us to make use of its form by including several types of the global constraints related to classes. In particular we consider all linear constraints of indicator variables of the following form

$$\sum_{j \in \mathcal{V}} \sum_{p=1}^{P} w_{jp}^m y_{jp} = c^m, \quad m = 1, \ldots, M \tag{12}$$

$$\sum_{j \in \mathcal{V}} \sum_{p=1}^{P} v_{jp}^k y_{jp} \leq d^k, \quad k = 1, \ldots, K. \tag{13}$$

---
[3]If it is not possible to make an update according to these rules than we obtain a strong agreement situation.

Then $p^{th}$ subproblem takes the form

$$\Phi_p(Y_p, \Lambda, M, K) = E_p(Y_p) + \sum_{j \in \mathcal{V}} \lambda_j y_{jp} + \sum_{m=1}^{M} \mu_m \sum_{j \in \mathcal{V}} w_{jp}^m y_{jp} + \sum_{k=1}^{K} \kappa_k \sum_{j \in \mathcal{V}} v_{jp}^k y_{jp} \to \min_{Y_p \in \mathcal{L}}$$

The subproblem is still submodular. The maximization of

$$\min_{Y \in \mathcal{L}} L(Y, \Lambda, M, K) = \sum_{p=1}^{P} \min_{Y_p} \Phi_p(Y_p, \Lambda, M, K) - \sum_{j \in \mathcal{V}} \lambda_j - \sum_{m=1}^{M} \mu_m c^m - \sum_{k=1}^{K} \kappa_k d^k$$

w.r.t. Lagrange multipliers $\Lambda$, $M$, and $K \geq \vec{0}$ can be done e.g. via subgradient ascent since the function is piecewise linear and concave. Note that each global constraint adds just one additional variable in the Lagrangian.

Here are several examples of global linear constraints:

$$\sum_{j \in \mathcal{V}} y_{jp} = c, \quad \text{Strict class size constraints} \tag{14}$$

$$\sum_{j \in \mathcal{V}} y_{jp} \in [c_1, c_2], \quad \text{Soft class size constraints} \tag{15}$$

$$\sum_{j \in \mathcal{V}} y_{jp} I_j = \sum_{j \in \mathcal{V}} y_{jq} I_j, \quad \text{Flux equalities} \tag{16}$$

$$\begin{cases} \sum_{j \in \mathcal{V}} y_{jp} I_j = \vec{\mu} \sum_{i \in \mathcal{V}} y_{ip}, & \text{Color mean/} \\ \sum_{j \in \mathcal{V}} y_{jp} (I_j - \vec{\mu})(I_j - \vec{\mu})^T = \sum_{j \in \mathcal{V}} y_{ip} \Sigma, & \text{variance.} \end{cases} \tag{17}$$

Here $I_j$ is an observable scalar or vector value associated with node, e.g. it can be the color of an image or the intensity of a grayscale image.

Such global linear statistics cannot be used in the state-of-the-art $\alpha$-expansion [5] method since it does not provide a lower bound for energy and their inclusion in TRW-like algorithms results in the following problem

$$L_{TRW}(Y, \tilde{\Theta}) - \sum_m \mu_m c^m - \sum_k \kappa_k d^k \to \max_{\tilde{\Theta}, M, K} \tag{18}$$

$$s.t. \sum_T \rho^T \tilde{\vartheta}^T \equiv \vartheta$$

$$\vartheta_{jp} = \theta_{jp} + \sum_m \mu_m w_{jp}^m + \sum_k \kappa_k v_{jp}^k$$

$$\vartheta_{ij,pq} = \theta_{ij,pq}, \quad \kappa_k \geq 0.$$

Here $T$ indexes the set of subgraphs that are used in TRW decomposition, $L(Y, \tilde{\Theta})$ is sum of optimal energies computed on each of the subgraphs, and $\equiv$ defines reparameterization relation. The straightforward way to solve the problem leads to nested iterative process. On the inner loop we maximize $L_{TRW}$ given $\vartheta$ and on the outer loop we update $M$ and $K$ by making subgradient ascent. We will refer to this algorithm as GTRW. One could use a linear programming relaxation of (18) but then it's not clear how to solve it efficiently in the case of general linear constraints such as (14)–(17) (invoking a general-purpose LP solver may be too slow in practice).

Another promising option of SMD is its ability to take into account some shape constraints. In particular we may establish all shape constraints that can be achieved by binary graphcut algorithm, e.g. star-shape prior [27] and "boundary position" prior [7] for any number of classes.

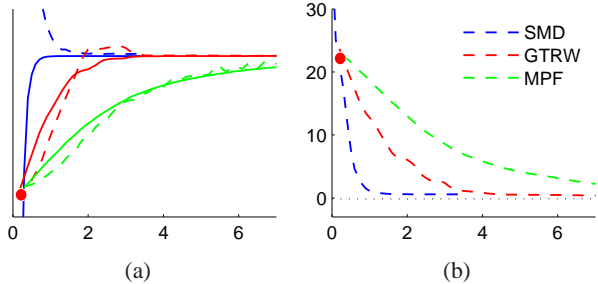

Figure 2: Results on a synthetic test of SMD, GTRW, and MPF. (a): solid lines correspond to lower bounds of different methods, dashed lines show energy of primal solution (that violates global hard constraints) at each iteration; (b): the total constraint violation of primal solution (measured in percentage of pixels that has to be recolored to make constraints consistent). On both plots horizontal axes show time measured in seconds. The red circles on both plots show the result of TRW-S algorithm as a "starting point" of GTRW and MPF.

## 7 Experiments

**Synthetic problems.** We compare the convergence rates of different methods on a number of artificially generated problems with linear constraints. Similarly to [13] as a synthetic setting we take a 10-label problem on a $50 \times 50$ grid graph with 4-neighborhood system. Unary potentials are generated as independent gaussians: $\theta_{ip} \sim \mathcal{N}(0, 1)$. Parameters $C_{ij,p} = C_{ij}$ of pairwise potentials are generated as an absolute value of $\mathcal{N}(0, 0.5)$. As global constraints we've chosen strict class size constraints (14). Class sizes are deliberately set to be significantly different from the global minimum of unconstrained energy (1). Particularly, for class $p$ we set its size to be $|\mathcal{V}|p/\sum_{q=1}^{P} q$.

Besides SMD we evaluate the performance of GTRW with TRW-S [13] as an inner iterative process. The third method is based on MPF framework proposed in [32]. Let $\Phi_L(\Pi)$ be the lower bound of energy (1) with unary potentials $\vartheta_{jp} = \theta_{jp} - \pi_{jp}$ and pairwise terms $\vartheta_{ij,pq} = \theta_{ij,pq}$; $\Phi_G(\Pi) = \min_Y \langle \Pi, Y \rangle$ s.t. $\sum_j y_{jp} = c_p \geq 0$, $\sum_p c_p = |\mathcal{V}|$, $Y \in \mathcal{Q} \cap \mathcal{G}$. Then we can maximize $\Phi_L(\Pi) + \Phi_G(\Pi)$ w.r.t. $\Pi$ via subgradient ascent. We computed $\Phi_L(\Pi)$ using TRW-S algorithm and $\Phi_G(\Pi)$ using simplex method for transportation problem.

Figure 2 shows the convergence of SMD, GTRW, and MPF averaged over 50 randomly generated problems. For MPF we do not consider the time required to solve the transportation problem $\Phi_G(\Pi)$ since it is highly dependent on the effectiveness of the implementation. Plot (a) shows lower bounds and energies of primal solutions[4] of all three methods. (b) shows the reduction of the total constraint violation of the current solution. Note, that energy of primal solution can be lower than the lower bound since it, generally speaking, does not satisfy constraints. Moreover, in GTRW and MPF the energy of the current solution increases since both methods start from energy minimum found by TRW-S and try to make the solution consistent with constraints. Based on figure 2 we conclude that all three methods converge to the same point but in SMD lower bound converges faster and hence primal solution with both low energy and small constraint violation can be found.

**Magnetogram segmentation.** We demonstrate that SMD can be used for implying flux constraints. Figure 3a, b show photos of the Sun in ultraviolet and magnetogram specters. In the regions of increased sun activity (Figure 3c) the amplitude of magnetic field is significantly larger (large positive magnetic fields are shown in white while large negative fields are shown in black). The greater is the deviation of pixel intensity from the gray level, the larger is the amplitude of magtetic field. The segmentation of sun's surface into quiet regions and the regions with large positive and negative magnetic fields and the analysis of their mutual allocation is important for further forecast of sun flares. It is known that for stable sunspots the total positive flux approximately equals the total negative flux in some vicinity of a sunspot. This constraint is a particular case of (16) and hence can be taken into account by SMD. Figures 3d–e shows how the segmentation changed after we have added flux constraints to our energy. Note that the major changes happened in the regions where the amplitude of magnetic field was relatively low and therefore those regions could be treated as quiet ones.

---
[4] The choice of primal solution is always an heuristics in dual decomposition methods. In SMD all pixels with conflicting labels in the same connected component were assigned to the same randomly selected label from the set of conflicting labels. In GTRW primal solution was chosen as TRW-S primal solution. In MPF primal solution was chosen as a solution of local subproblem $\Phi_L(\Pi)$ solved by TRW-S (solution of global subproblem $\Phi_G(\Pi)$ always satisfies global constraints but typically has much higher energy).

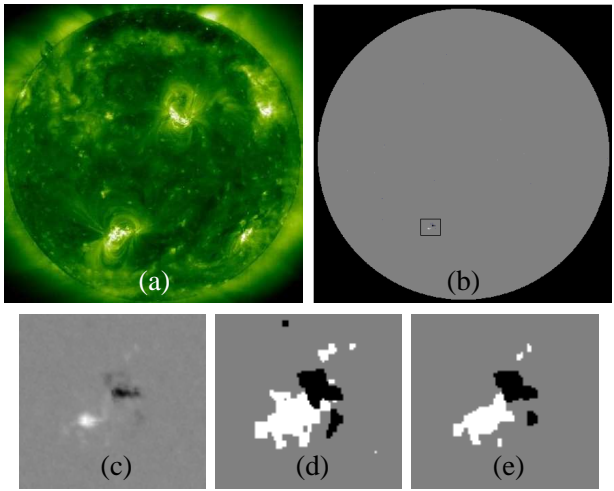

Figure 3: Results on a magnetogram test. (a) – the picture of the Sun at 195 angstrom wavelength in the extreme ultraviolet; (b) – corresponding magnetogram: white regions correspond to the areas with large positive magnetic field, black regions – to large negative magnetic field, gray regions – to relatively inactive areas; (c) – (b) – the increased region of high activity; (d) – the result of $\alpha$-expansion without global constraints: positive flux = 83794, negative flux = -71021, diff = 12773; (e) – the result of SMD with flux equality constraint: positive flux = 70972, negative flux = -67341, diff = 3630.

**Image segmentation.** SMD framework gives an opportunity to take into account any constraints that can be worked out with a single binary graphcut. As an example we present our results of including star-shape prior [27] into SMD framework for image segmentation. We work in a framework that can be viewed as a multilabel version of [3], i.e. user defines a small seed region for each class which we use to collect color statistics and to establish the center for star-shape prior.

As unary potentials we use standard $-\log P(I_i \mid x_i)$ pixel. Color statistics is collected from a small seed region that is provided by user for each class. As pairwise potentials we use generalized Potts Model $C_{ij} = a_1 + a_2 \exp\left(-\frac{(I_i - I_j)^2}{2\sigma^2}\right)$. We use $a_1 = 2, a_2 = 20$; for each edge we set $\sigma$ to be the absolute average intensity difference in the box corresponding pixels. The box size is set to be 20 by 20.

A star-shape prior is a compact shape priors that given a center $c$ makes an object to be star-shaped around the center, i.e. on any ray starting at $c$ the "object" label 1 cannot appear after the "background" label 0. It can be written down via pairwise potentials in the following way:

$$S_{ij} = \begin{cases} 0, & \text{if } f_i = f_j, \\ \infty, & \text{if } f_i = 1 \text{ and } f_j = 0, \\ \beta, & \text{if } f_i = 0 \text{ and } f_j = 1, \end{cases}$$

where $j$ is between $c$ and $i$. Parameter $\beta$ corresponds to the "ballooning force" and is set to 0 in our experiments. For more details on star-shape priors see [27].

Figure 4 presents our results. Note that TRW-S and $\alpha$-expansion converge to similar almost optimal solution but it has poor quality unless we take into account global constraints. This effect is a result of mixed color statistics in unary terms.

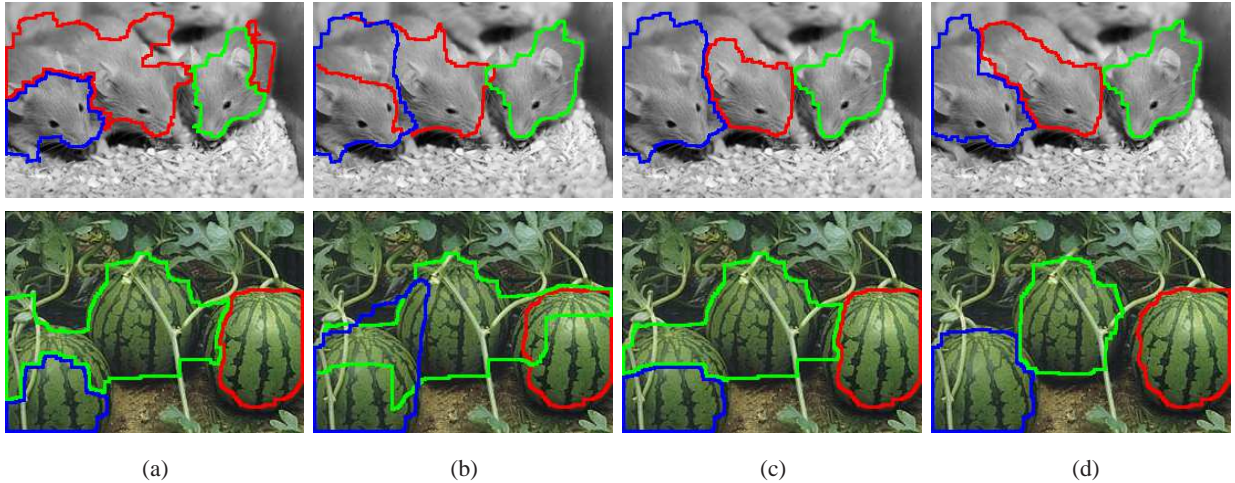

(a) (b) (c) (d)

Figure 4: Results for the image segmentation problem. (a) – TRW-S and $\alpha$-expansion results; (b) – the results of independent usage of binary graphcut with star-shape prior for each object; (c) – the result of SMD with star-shape prior; (d) – the result of SMD with both star-shape prior and class size equality constraints (all classes except background should have equal sizes).

# 8 Discussion and Future Work

In the paper we presented a novel approach to approximate Bayesian inference in associative MRFs with cycles based on the decomposition of initial NP-hard problem into a number of submodular problems. The advantage of such approach is the fact that each subproblem is related to a specific class. Hence any global statistics of classes can be added directly to the subproblems rather than by organizing iterative process which requires full MAP-inference on each iteration. Some questions still remain open:

- Is it possible to derive an efficient algorithm for min-marginal averaging?
- How the value of the gap between minimax and maximin depends on the number of global constraints?
- Are there any partial label optimality guarantee for partially coincident subproblems solutions?

Two directions of further research work on SMD should be mentioned. We plan to adapt the SMD framework for a wider class of pairwise terms (e.g. for the cases when pairwise terms define semi-metrics in the space of class labels) and to add more elaborate shape priors as global terms.

# Appendix

## A  Proof of theorem 1

In this section we prove that the SMD decomposition is equivalent to the LP relaxation of Kleinberg and Tardos (9):

$$\max_{\Lambda} \min_{Y \in \mathcal{L}} L(Y, \Lambda) = \min_{X} \sum_{j \in \mathcal{V}} \sum_{p=1}^{P} \theta_{jp} x_{jp} + \frac{1}{2} \sum_{(i,j) \in \vec{\mathcal{E}}} \sum_{p=1}^{P} C_{ij,p} x_{ij,p}$$

$$\text{s.t.} \quad \sum_{p=1}^{P} x_{jp} = 1 \qquad \forall j \in \mathcal{V}$$

$$x_{ij,p} \geq x_{ip} - x_{jp}, \qquad \forall (i \to j) \in \vec{\mathcal{E}}, p \in \{1, \ldots, P\}$$

$$x_{ij,p} \geq 0, \; x_{jp} \geq 0$$

Here we define $(\mathcal{V}, \vec{\mathcal{E}})$ to be the directed graph corresponding to the undirected graph $(\mathcal{V}, \mathcal{E})$, i.e. $\vec{\mathcal{E}} = \{(i \to j), (j \to i)\} \mid (i,j) \in \mathcal{E}\}$. We can rewrite the Langrangian (6) as follows:

$$L(Y, \Lambda) = \sum_{j \in \mathcal{V}} \sum_{p=1}^{P} \theta_{jp} y_{jp} + \frac{1}{2} \sum_{(i,j) \in \mathcal{E}} \sum_{p=1}^{P} C_{ij,p} |y_{ip} - y_{jp}| + \sum_{j \in V} \lambda_j \left( \sum_{p=1}^{P} y_{jp} - 1 \right)$$

The minimum of the Lagrangian over $Y \in \mathcal{L}$ equals

$$\Phi(\Lambda) \equiv \min_{Y \in \mathcal{L}} L(Y, \Lambda)$$

$$\stackrel{(1)}{=} \min \quad \sum_{j \in \mathcal{V}} \sum_{p=1}^{P} \theta_{jp} y_{jp} + \frac{1}{2} \sum_{(i \to j) \in \vec{\mathcal{E}}} \sum_{p=1}^{P} C_{ij,p} y_{ij,p} + \sum_{j \in \mathcal{V}} \lambda_j \left( \sum_{p=1}^{P} y_{jp} - 1 \right)$$

$$\text{s.t.} \quad y_{jp} \leq 1 \qquad \forall j \in \mathcal{V}, p \in \{1, \ldots, P\}$$

$$y_{ij,p} \geq y_{ip} - y_{jp} \qquad \forall (i \to j) \in \vec{\mathcal{E}}, p \in \{1, \ldots, P\}$$

$$y_{jp} \geq 0, \; y_{ij,p} \geq 0$$

$$\stackrel{(2)}{=} \max \quad -\sum_{j \in \mathcal{V}} \sum_{p=1}^{P} \mu_{jp} - \sum_{j \in \mathcal{V}} \lambda_j$$

$$\text{s.t.} \quad -\mu_{jp} - \sum_{(i \to j)} \mu_{ij,p} + \sum_{(j \to i)} \mu_{ji,p} \leq \theta_{jp} + \lambda_j \qquad \forall j \in \mathcal{V}, p \in \{1, \ldots, P\}$$

$$\mu_{ij,p} \leq \frac{1}{2} C_{ij,p} \qquad \forall (i \to j), p$$

$$\mu_{jp} \geq 0, \; \mu_{ij,p} \geq 0$$

where (1) uses a standard relaxation of the $s$-$t$ min cut problem (which is known to have no duality gap), and (2) uses LP duality. The maximization problem $\max_\Lambda \Phi(\Lambda)$ is now a linear program; its dual is

$$
\begin{aligned}
\Phi^* &= \max_\Lambda \Phi(\Lambda) \\
&= \min \sum_{j \in \mathcal{V}} \sum_{p=1}^{P} \theta_{jp} x_{jp} + \frac{1}{2} \sum_{(i \to j) \in \vec{\mathcal{E}}} \sum_{p=1}^{P} C_{ij,p} x_{ij,p} \\
\text{s.t.} \quad & x_{jp} \leq 1 & \forall j \in \mathcal{V}, p \in \{1, \ldots, P\} \\
& x_{ij,p} \geq x_{ip} - x_{jp} & \forall (i \to j) \in \vec{\mathcal{E}}, p \in \{1, \ldots, P\} \\
& \sum_{p=1}^{P} x_{jp} = 1 & \forall j \in \mathcal{V} \\
& x_{jp} \geq 0, \ x_{ij,p} \geq 0
\end{aligned}
$$

which is the LP of Kleinberg and Tardos.

# B  Proof of theorem 2

This section shows that the LP of Kleinberg and Tardos is equivalent to the Shlesinger LP in case of Potts model.

Given two vectors $x, y \in [0,1]^P$ with $\sum_{p=1}^{P} x_p = \sum_{p=1}^{P} y_p = 1$ and non-negative constants $C_p$, $p = 1, \ldots, P$, define

$$f(x, y) = \min \sum_{p=1}^{P} \sum_{q=1}^{P} d_{pq} z_{pq} \qquad (19)$$

$$\text{s.t.} \quad \sum_{q=1}^{P} z_{pq} = x_p \qquad (20)$$

$$\sum_{p=1}^{P} z_{pq} = y_q \qquad (21)$$

$$z_{pq} \geq 0 \qquad (22)$$

where $d_{pq} = 0$ if $p = q$, and $d_{pq} = \frac{C_p + C_q}{2}$ otherwise. It suffices to show that $f(x, y) = \frac{1}{2} \sum_{p=1}^{P} C_p |x_p - y_p|$.

**Lemma 2** *Denote $z_{pq}^\circ = \min\{x_p, y_q\}$. There exists vector $z \in [0,1]^{P \times P}$ with $z_{pp} = z_{pp}^\circ$ satisfying constraints (20), (21), (22).*

**Proof.** Consider the following algorithm for constructing $z$:

S0 Initialize: $z_{pp} = z_{pp}^\circ$, $z_{pq} = 0$ for $p \neq q$.

S1 While there is a pair of labels $(p, q)$ with $x_p > \sum_{q'=1}^{P} z_{pq'}$ and $y_p > \sum_{p'=1}^{P} z_{p'q}$, do the following:

- compute $\delta = \min \left\{ x_p - \sum_{q'} z_{pq'}, y_p - \sum_{p'=1}^{P} z_{p'q} \right\}$
- set $z_{pq} := z_{pq} + \delta$

Clearly, the algorithm maintains the following invariants:

$$x_p \geq \sum_{q'=1}^{P} z_{pq'} \qquad y_q \geq \sum_{p'=1}^{P} z_{p'q}$$

Furthermore, components of vector $z$ never decrease, and after the update for $(p,q)$ we have either $x_p = \sum_{q'=1}^{P} z_{pq'}$ or $y_q = \sum_{p'=1}^{P} z_{p'q}$. Thus, the algorithm terminates in a finite number of steps. It can be seen that upon termination

$$x_p = \sum_{q'=1}^{P} z_{pq'} \qquad y_q = \sum_{p'=1}^{P} z_{p'q}$$

for all $p$ and $q$. Indeed, suppose, for example, that upon termination $x_p > \sum_{q'=1}^{P} z_{pq'}$ for some $p$, then there must hold $y_q = \sum_{p'=1}^{P} z_{p'q}$ for all $q$, otherwise the algorithm wouldn't have terminated. Then

$$1 = \sum_{p=1}^{P} x_p > \sum_{p=1}^{P} \sum_{q'=1}^{P} z_{pq'} = \sum_{q=1}^{P} \sum_{p'=1}^{P} z_{p'q} = \sum_q y_q = 1$$

- a contradiction.

It remains to note that during the algorithm $z_{rr} = z_{rr}^\circ$ for all $r \in \{1, \ldots, P\}$ since either $x_r = \sum_{q=1}^{P} z_{rq}$ or $y_r = \sum_{p=1}^{P} z_{pr}$. ∎

Let $z$ be the vector constructed in the lemma, then

$$\begin{aligned}
\sum_{p=1}^{P} \sum_{q=1}^{P} d_{pq} z_{pq} &= \frac{1}{2} \left( \sum_{p=1}^{P} C_p x_p + \sum_{q=1}^{P} C_q y_q - 2 \sum_{r=1}^{P} C_r z_{rr} \right) \\
&= \frac{1}{2} \left( \sum_{r=1}^{P} C_r \min\{x_r, y_r\} + \sum_{r=1}^{P} C_r \max\{x_r, y_r\} - 2 \sum_{r=1}^{P} C_r \min\{x_r, y_r\} \right) \\
&= \frac{1}{2} \left( \sum_{r=1}^{P} C_r \max\{x_r, y_r\} - \sum_{r=1}^{P} C_r \min\{x_r, y_r\} \right) = \frac{1}{2} \sum_{r=1}^{P} C_r |y_r - x_r|
\end{aligned}$$

Let $z'$ be another vector satisfying (20), (21), (22). We must have $z'_{rr} \leq z_{rr} = \min\{x_r, y_r\}$, therefore

$$\begin{aligned}
\sum_{p=1}^{P} \sum_{q=1}^{P} d_{pq} z'_{pq} &= \frac{1}{2} \left( \sum_{p=1}^{P} C_p x_p + \sum_{q=1}^{P} C_q y_q - 2 \sum_{r=1}^{P} C_r z'_{rr} \right) \\
&\geq \frac{1}{2} \left( \sum_{p=1}^{P} C_p x_p + \sum_{q=1}^{P} C_q y_q - 2 \sum_{r=1}^{P} C_r z_{rr} \right) = \sum_{p=1}^{P} \sum_{q=1}^{P} d_{pq} z_{pq}
\end{aligned}$$

This shows that $f(x, y) = \frac{1}{2} \sum_{r=1}^{P} C_r |y_r - x_r|$, as claimed.

## C  Proof of lemma 1

Since $\min_{Y \in \mathcal{L}} L(Y, \Lambda)$ is a piecewise linear function three cases are possible at its maximum (see fig. 2). In the first case there is a plateau on the top of the polytope. If subgradient consists of only one vector then the function is differentiable at the point $\Lambda^0$, i.e. $\Lambda^0$ is located somewhere *inside* the horizontal face that is defined by $\arg\min L(Y, \Lambda^0) = Y^0$. But this means that $L(Y^0, \Lambda)$ does not depend on $\Lambda$, i.e. $Y^0 \in \mathcal{G}$. For all $Y \in \mathcal{G}$ holds $L(Y, \Lambda) = E(Y)$ and $\min_{Y \in \mathcal{L} \cap \mathcal{G}} L(Y, \Lambda) = \min_{Y \in \mathcal{L} \cap \mathcal{G}} E(Y)$.

Consider the second case when there is no plateau but there exists a horizontal plane which passes through the point $(\Lambda^0, L(Y^0, \Lambda^0))$. Let $Y^1$ be the value of $Y \in \mathcal{L}$ which corresponds to that plane. Since the plane is horizontal $Y^1 \in \mathcal{G}$ as well. Then it holds
$$L(Y^0, \Lambda^0) = L(Y^1, \Lambda^0) = \min_{Y \in \mathcal{L} \cap \mathcal{G}} L(Y, \Lambda^0) = \min_{Y \in \mathcal{L} \cap \mathcal{G}} E(Y).$$

Now assume that $L(Y^0, \Lambda^0) = \min_{Y \in \mathcal{L} \cap \mathcal{G}} E(Y) = E(Y^1)$. On the other hand $\min_{Y \in \mathcal{L} \cap \mathcal{G}} E(Y) = E(Y^1) = L(Y^1, \Lambda^0)$. Hence $L(Y^0, \Lambda^0) = L(Y^1, \Lambda^0)$. From that it follows that there is a plane which corresponds to $Y^1$ passing through the point $(\Lambda^0, L(Y^0, \Lambda^0))$. Since $Y^1 \in \mathcal{G}$ the plane is horizontal and we have either the first or the second case.

Finally there can be the third case when there's no horizontal plane at maximin point. In this case a non-zero duality gap takes place and we can only get the lower bound for optimal value of energy $E(Y)$.

## D  Proof of theorem 9

In this section we prove that min-marginal averaging process defined by (11) if monotonic and satisfies weak agreement condition at the fixed point.

**Lemma 3** *Every submodular function*
$$E(\vec{x}) = \sum_{(j \in \mathcal{V})} \theta_j(x_j) + \sum_{(i,j) \in \mathcal{E}} \theta_{ij}(x_i, x_j)$$

*where $x_j \in \{0, 1\}$ can be reparameterized in such a way*
$$E(\vec{x}) = \sum_{(j \in \mathcal{V})} \hat{\theta}_j(x_j) + \sum_{(i,j) \in \mathcal{E}} \hat{\theta}_{ij}(x_i, x_j) + Const,$$

$$\hat{\theta}_j(k) = \min_{\vec{x}} E(\vec{x} | x_j = k) - E_0, \tag{23}$$

$$\hat{\theta}_i(l) + \hat{\theta}_{ij}(l, k) + \hat{\theta}_j(k) = \min_{\vec{x}} E(\vec{x} | x_i = l, x_j = k) - E_0, \tag{24}$$

*where $E_0 = \min_{\vec{x}} E(\vec{x})$.*

**Proof.** We perform tree-based decomposition and make use of the fact that TRW converges to optimal value of energy (strong tree agreement) for submodular functions.

Consider a decomposition of MRF into a set of trees $\mathcal{T}$ such that each edge $(i, j) \in \mathcal{E}$ is covered by only one tree. It can be shown that each tree $T$ can be reparameterized the following form [28, theorem 1]:

$$\hat{\theta}_j^T(k) = \min_{\vec{x}} E^T(\vec{x} | x_j = k) - E_T, \tag{25}$$

$$\hat{\theta}_i^T(l) + \hat{\theta}_{ij}^T(l, k) + \hat{\theta}_j^T(k) = \min_{\vec{x}} E^T(\vec{x} | x_i = l, x_j = k) - E_T, \tag{26}$$

where $E_T = \min_{\vec{x}} E^T(\vec{x})$. Using the fact that for submodular functions we have strong tree agreement [14, theorem 3] we may write

$$E_0 = \min_{\vec{x}} E(\vec{x}) = \sum_{T \in \mathcal{T}} \min_{\vec{x}} E^T(\vec{x}) = \sum_{T \in \mathcal{T}} E_T.$$

The same is true for conditional minimums since they are minimums of submodular functions $\min_{\vec{x}} E(\vec{x} | x_j = k)$ as well:

$$\min_{\vec{x}} E(\vec{x} | x_j = k) = \sum_{T \in \mathcal{T}} \min E^T(\vec{x} | x_j = k) = \sum_{T \in \mathcal{T}} \left( \hat{\theta}_j^T(k) + E_T \right) = \sum_{T : x_j \in \mathcal{V}_T} \hat{\theta}_j^T(k) + E_0.$$

Each $\{\hat{\theta}^T\}$ defines energy $E^T$, hence their sum defines the initial energy $E$.

$$\hat{\theta}_j(k) = \sum_{T: x_j \in \mathcal{V}_T} \hat{\theta}_j^T(k) = \min_{\vec{x}} E(\vec{x}|x_j = k) - E_0.$$

Similar reasoning is valid for the edges

$$\min_{\vec{x}} E(\vec{x}|x_i = l, x_j = k) = \sum_{T \in \mathcal{T}} \min_{\vec{x}} E^T(\vec{x}|x_i = l, x_j = k) = \sum_{T: x_i \in \mathcal{V}_T,\, x_j \notin \mathcal{V}_T} \min_{\vec{x}} E^T(\vec{x}|x_i = l) +$$
$$+ \sum_{T: x_j \in \mathcal{V}_T,\, x_i \notin \mathcal{V}_T} \min_{\vec{x}} E^T(\vec{x}|x_j = k) + \min_{\vec{x}} E^{T_{ij}}(\vec{x}|x_i = l, x_j = k) =$$
$$= \sum_{T: x_i \in \mathcal{V}_T,\, x_j \notin \mathcal{V}_T} \left(\hat{\theta}_i^T(l) + E_T\right) + \sum_{T: x_j \in \mathcal{V}_T,\, x_i \notin \mathcal{V}_T} \left(\hat{\theta}_j^T(k) + E_T\right) + \hat{\theta}_i^{T_{ij}}(l) + \hat{\theta}_{ij}^{T_{ij}}(l, k) + \hat{\theta}_j^{T_{ij}}(k) + E_{T_{ij}} =$$
$$= \sum_{T: x_i \in \mathcal{V}_T} \hat{\theta}_i^T(l) + \sum_{T: x_j \in \mathcal{V}_T} \hat{\theta}_j^T(k) + \hat{\theta}_{ij}^{T_{ij}}(l, k) + E_0 = \hat{\theta}_i(l) + \hat{\theta}_j(k) + \hat{\theta}_{ij}(l, k) + E_0,$$

where $T_{ij}$ is the tree that covers edge $(i, j)$. ∎

Now we prove theorem 9. Consider the change of a single $\lambda_j$ with all other $\lambda$'s unchanged. According to the previous lemma we can use the reparameterization defined by (23), (24). Let $s = \arg\max_q \left[\hat{\theta}_{jq}(0) - \hat{\theta}_{jq}(1)\right]$. Then

$$\Delta \lambda_j = \hat{\theta}_{js}(0) - \hat{\theta}_{js}(1).$$

If $\Delta \lambda_j \geq 0$ then

$$\min_{Y_p} \Phi_p(Y_p, \Lambda^{new}) \geq \min_{Y_p} \Phi_p(Y_p, \Lambda^{old}), \quad \forall p \neq s$$

and

$$\min_{Y_s} \Phi_p(Y_s, \Lambda^{new}) = \min_{Y_s} \Phi_p(Y_s, \Lambda^{old}) + \Delta \lambda_j.$$

Last equation follows from the fact that $\hat{\theta}_{js}(1) = 0$ and $Z_{js}(\Lambda^{new}) = \{0, 1\}$. Using the fact that

$$L(Y, \Lambda) = \sum_{p=1}^{P} \min_{Y_p} \Phi_p(Y_p, \Lambda) - \sum_{j \in \mathcal{V}} \lambda_j$$

we obtain

$$L(Y, \Lambda^{new}) \geq L(Y, \Lambda^{old}) + \Delta \lambda_j - \Delta \lambda_j = L(Y, \Lambda^{old}).$$

Now consider the case when $\Delta \lambda_j < 0$. Hence it follows that all $\hat{\theta}_{jr}(0) = 0$ and consequently $\hat{\theta}_{js}(1) = \min_r \hat{\theta}_{jr}(1)$. The decrease of $\lambda_j$ by $\Delta \lambda_j = -\hat{\theta}_{js}(1)$ still keeps $\{0\} \in Z_{jp}$ for all $p$, i.e.

$$\min \Phi_p(Y_p, \Lambda^{new}) = \min \Phi_p(Y_p, \Lambda^{old}), \quad \forall p.$$

The lower bound increases because of the last term

$$L(Y, \Lambda^{new}) = L(Y, \Lambda^{old}) - \Delta \lambda_j > L(Y, \Lambda^{old}).$$

Now we prove that the fixed point of the iterative process always satisfies weak agreement condition. Let all $\Delta \lambda_j = 0$ for some value of $\Lambda^0$. Since $\max_q \left[\hat{\theta}_{jq}(0) - \hat{\theta}_{jq}(1)\right] = 0$ for all vertices $j$, we observe that $1 \in Z_{js}$ and $\forall p: 0 \in Z_{jp}$, which implies weak agreement condition.

# References

[1] D. Batra, A. C. Gallagher, D. Parikh, and T. Chen. Beyond trees: MRF inference via outer-planar decomposition. In *CVPR*, 2010.

[2] D. Bertsekas. *Nonlinear Programming*. Athena Scientific, 1999.

[3] Y. Boykov and G. Funka-Lea. Graph Cuts and Efficient N-D Image Segmentation. *IJCV*, 70(2):109–131, 2006.

[4] Y. Boykov and V. Kolmogorov. An Experimental Comparison of Min-Cut/Max-Flow Algorithms for Energy Minimization in Vision. *PAMI*, 26(9):1124–1137, 2004.

[5] Y. Boykov, O. Veksler, and R. Zabih. Fast Approximate Energy Minimization via Graph Cuts. *PAMI*, 23(11):1222–1239, 2001.

[6] J. Darbon. Global optimization for first order Markov random fields with submodular priors. *Discrete Applied Mathematics*, 157(16):3412–3423, 2009.

[7] D. Freedman and T. Zhang. Interactive Graph Cut Based Segmentation With Shape Priors. In *CVPR*, 2005.

[8] D. Greig, B. Porteous, and A. Seheult. Exact maximum a posteriori estimation for binary images. *Journal of the Royal Statistical Society, Series B*, 51(2):271–279, 1989.

[9] H. Ishikawa. Exact Optimization for Markov Random Fields with Convex Priors. *PAMI*, 25(10):1333–1336, 2003.

[10] J. M. Kleinberg and É. Tardos. Approximation algorithms for classification problems with pairwise relationships: metric labeling and Markov random fields. *J. ACM*, 49(5):616–639, 2002.

[11] P. Kohli and P. Torr. Measuring Uncertainty in Graph Cut Solutions. In *ECCV*, 2006.

[12] P. Kohli and P. Torr. Dynamic Graph Cuts for Efficient Inference in Markov Random Fields. *PAMI*, 29(12):2079–1088, 2007.

[13] V. Kolmogorov. Convergent Tree-Reweighted Message Passing for Energy Minimization. *PAMI*, 28(10):1568–1583, 2006.

[14] V. Kolmogorov and M. Wainwright. On the Optimality of Tree-reweighted Max-product Message Passing. In *UAI*, 2005.

[15] V. Kolmogorov and R. Zabih. What Energy Functions Can Be Optimized via Graph Cuts. *PAMI*, 26(2):147–159, 2004.

[16] N. Komodakis and N. Paragios. Beyond loose LP-relaxations: Optimizing MRFs by repairing cycles. In *ECCV*, 2008.

[17] N. Komodakis, N. Paragios, and G. Tziritas. MRF Energy Minimization and Beyond via Dual Decomposition. *PAMI*, 33(3):531–552, 2010.

[18] N. Komodakis and G. Tziritas. Approximate Labeling via Graph-Cuts Based on Linear Programming. *PAMI*, 29(8):1436–1453, 2007.

[19] S. L. Lauritzen. *Graphical models*. Clarendon Press, 1996.

[20] Y. Lim, K. Jung, and P. Kohli. Energy Minimization Under Constraints on Label Counts. In *ECCV*, 2010.

[21] J. Pearl. *Probabilistic Reasoning in Intelligent Systems: Networks of Plausible Inference*. Morgan Kaufman, San Francisco, 1988.

[22] P. Ravikumar and J. Lafferty. Quadratic Programming Relaxations for Metric Labeling and Markov Random Fields MAP Estimation. In *ICML*, 2006.

[23] N. N. Schraudolph and D. Kamenetsky. Efficient Exact Inference in Planar Ising Models. In *NIPS*, 2008.

[24] D. Sontag, A. Globerson, and T. Jaakkola. Introduction to dual decomposition for inference. In S. Sra, S. Nowozin, and S. J. Wright, editors, *Optimization for Machine Learning*. MIT Press, 2011.

[25] D. Sontag, T. Meltzer, A. Globerson, Y. Weiss, and T. Jaakkola. Tightening LP relaxations for MAP using message-passing. In *UAI*, 2008.

[26] B. Taskar, V. Chatalbashev, and D. Koller. Learning Associative Markov Networks. In *ICML*, 2004.

[27] O. Veksler. Star Shape Prior for Graph-Cut Image Segmentation. In *ECCV*, 2008.

[28] M. J. Wainwright, T. S. Jaakkola, and A. S. Willsky. Tree consistency and bounds on the performance of the max-product algorithm and its generalizations. *Statistics and Computing*, 14(2):143–166, 2004.

[29] M. J. Wainwright, T. S. Jaakkola, and A. S. Willsky. MAP estimation via agreement on trees: message-passing and linear programming. *IEEE Transactions on Information Theory*, 51(11):3697–3717, 2005.

[30] T. Werner. A Linear Programming Approach to Max-sum Problem: A Review. *PAMI*, 29(7):1165–1179, 2007.

[31] T. Werner. High-arity Interactions, Polyhedral Relaxations, and Cutting Plane Algorithm for Soft Constraint Optimisation (MAP-MRF). In *CVPR*, 2008.

[32] O. J. Woodford, C. Rother, and V. Kolmogorov. A Global Perspective on MAP Inference for Low-Level Vision. In *ICCV*, 2009.